\documentclass[a4paper] {article}

\usepackage{amsmath}
\usepackage{color}
\newcommand{\ra}{\rightarrow}

\newcommand{\N}{\mathcal{N}}

\usepackage{latexsym}
\usepackage{graphicx}

\makeatletter
\renewcommand\section{\@startsection{section}{1}{\z@}%
                                  {-3.5ex \@plus -1ex \@minus -.2ex}%
                                  {2.3ex \@plus.2ex}%
                                  {\normalfont\large\bfseries}}
\makeatother

\begin{document}

\begin{center}
{\fontsize{14}{17}\selectfont
\textbf{  Reinforcement Learning for Matrix \\ \ \\ Computations: PageRank as an Example}
}
\end{center}

\vspace{.5in}

\begin{center}
Vivek S.\ Borkar and Adwaitvedant S.\ Mathkar\\
\ \\
Department of Electrical Engineering,\\
Indian Institute of Technology, \\
Powai, Mumbai 400076,  India. \\
\{borkar.vs, mathkar.adwaitvedant\}@gmail.com
\end{center}

\vspace{.5in}

\noindent \textbf{Abstract.} Reinforcement learning has gained wide popularity as a technique for simulation-driven approximate dynamic programming. A less known aspect is that the very reasons that make it effective in dynamic programming can also be leveraged for using it for distributed schemes for certain matrix computations involving non-negative matrices. In this spirit, we propose a reinforcement learning algorithm for PageRank computation that is fashioned after analogous schemes for approximate dynamic programming. The algorithm has the advantage of ease of distributed implementation and more importantly, of being model-free, i.e., not dependent on any specific assumptions about  the transition probabilities in the random web-surfer model. We analyze its convergence and finite time behavior and present some supporting numerical experiments.\\

\noindent \textbf{Key words:} Reinforcement Learning, PageRank, Stochastic Approximation, Sample Complexity \\ \\

\section {Introduction}

Reinforcement learning has its roots in models of animal behavior \cite{Thorn}
and mathematical psychology \cite{Bush}, \cite{Estes}. The recent resurgence of interest in the field, however, is propelled by applications to artificial intelligence and control engineering. By now there are several textbook accounts of this development \cite{Bert} (Chapter 16), \cite{BerTsi}, \cite{Gosavi}, \cite{Powell}, \cite{Sutton}, \cite{Szep}. To put things in context, recall that methodologically, reinforcement learning sits somewhere in between supervised learning, which works with a reasonably accurate information regarding the performance gradient or something analogous (e.g., parameter tuning of neural networks), and unsupervised learning, which works without such explicit information (e.g., clustering). To be specific, supervised learning is usually based upon an optimization formulation such as minimizing an error measure, which calls for a higher quantum of information per iterate. Reinforcement learning on the other hand has to manage with signals somehow correlated with performance, but which fall short of the kind of information required for a typical supervised learning scheme. It then  makes simple incremental corrections based on these `suggestive though inexact' signals, usually with low per iterate computation. The latter aspect
has also made it a popular framework for models of bounded rationality in economics \cite{Sargent}.\\

Our interest is in its recent avatar as a scheme for simulation-based methodology for approximate dynamic programming for Markov decision processes which has found applications, among other things, in robotics \cite{Thrun}. These can be viewed as stochastic approximation counterparts of the classical iterative methods for solving dynamic programming equations, such as value and policy iteration. Stochastic approximation, introduced by Robbins and Monro \cite{Robbins} as an iterative scheme for finding the roots of a nonlinear function given its noisy measurements, is the basis of most adaptive schemes in control and signal processing. What it does in the present context is to replace a conditional average appearing on the right hand side of the  classical iterative schemes (or their variants) by an actual evaluation at a simulated transition according to the conditional distribution in question. It then makes an incremental move towards the resulting random quantity. That is, it takes a convex combination of the current value and the random right hand side, with a slowly decreasing weight on the latter. The averaging properties of stochastic approximation then ensure that asymptotically you see the same limiting behavior as the original scheme.\\

But there are other situations wherein one encounters iterations involving conditional averages. In fact, by pulling out row sums of a non-negative matrix into a diagonal matrix pre-multiplier, we can write it as a product of a diagonal matrix and a stochastic matrix. This allows us to cast iterations involving non-negative matrices as iterations involving averaging with respect to stochastic matrices, making them amenable to the above methodology. This opens up the possibility of using reinforcement learning schemes for distributed matrix computations of certain kind. Important instances are plain vanilla averaging  and estimation of the Perron-Frobenius eigenvectors \cite{snowwhite}. Reinforcement learning literature is replete with means of curtailing the curse of dimensionality, a hazard only too common in dynamic programming applications. This machinery then becomes available for such matrix computations. An important special case is the case of linear function approximation, wherein one approximates the desired vector by a weighted combination of a moderate number of basis vectors, and then updates these weights instead of the entire vector \cite{ISI}.\\

In the present article, we illustrate this methodology in the context of Google's PageRank, an eigenvector-based ranking scheme.
 It is primarily based on the stationary distribution $\pi$ of the `random web-surfer' Markov chain, equivalently, the normalized left Perron-Frobenius eigenvector of its transition probability matrix. This chain is defined on a directed graph  wherein each node $i$ is a web page. Let $\N(i) :=$ the set of nodes to which $i$ points. Let $d(i) := |\N(i)|$ and $N :=$ the total number of nodes. The chain moves from $i$ to $j \in \N(i)$ with a probability $(1 - c)\frac{1}{d(i)} + \frac{c}{N},$ and to any other node in the graph with probability $\frac{c}{N}$ where $c > 0$ is the `Google constant'. The latter renders it irreducible, ensuring a unique stationary distribution. An excellent account of the numerical techniques for computing $\pi$, essentially based on the `power method' and its variants, appears in \cite{LangMey1}, along with a brief historical account. See also \cite{LangMey2}. While a bulk of the work in this direction has been on efficient computations for the power method, there have also been alternative approaches, such as Markov Chain Monte Carlo \cite{Kostia1}, \cite{Kostia2}, optimization based methods \cite{Polyak}, and schemes based on stochastic approximation and/or gossip \cite{Ishii}, \cite{Nazin}, \cite{Zhao}.\\

Such `spectral ranking' techniques, made popular by the success of PageRank, are in fact quite old. See \cite{Vigna} for a historical survey. Evaluative exercises of this kind occur in other applications as well, such as reputation systems or popularity measures on social networks. In such applications (for that matter, in search), it is unclear whether the assumption that each $j \in \N(i)$ is equally important to $i$ is reasonable. Motivated by this, we propose a model-free scheme based on ideas from reinforcement learning. This idea has also been discussed in \cite{snowwhite}. The present scheme, however, differs in an essential way from \cite{snowwhite} in that whereas \cite{snowwhite} views PageRank as a special instance of the general problem of eigenvector estimation, we exploit the special structure of the random web-surfer model to simplify the problem to a simple linear scheme. This is very much in tune with some of the works cited above (notably \cite{Ishii}, \cite{Zhao}), but with a non-standard sample and update rule. The outcome is an algorithm that can run on accumulated traces of node-to-node interactions without requiring us to explicitly estimate the probabilities associated with these.\\

The next section describes our algorithm and its convergence analysis. Section 3 describes  finite time analysis and a variant of the basic scheme. Section 4 presents some numerical experiments. Section 5 concludes with some general observations.

\section{The Algorithm}
Let $P$ be an $N\times N$ stochastic matrix.
Define $\hat{P} := cP + \frac{1-c}{N}\begin{bmatrix} 1 & \cdots & 1  \\  \vdots & \ddots & \vdots  \\   1 & \cdots & 1 \end{bmatrix}$. Let $\pi$ denote the unique stationary probability distribution of $\hat{P}$. That is, for $\textbf{1} := [1, 1, \cdots, 1]^T$,

\begin{eqnarray*}
\pi &=& \pi\hat{P} \\
&=& \pi cP + \pi\frac{1-c}{N}\begin{bmatrix} 1 & \cdots & 1  \\  \vdots & \ddots & \vdots  \\   1 & \cdots & 1 \end{bmatrix} \\
&=& c\pi P + \frac{1-c}{N} \textbf{1}^{T} \\
\Rightarrow \pi(I - cP) &=& \frac{1-c}{N} \textbf{1}^{T} \\
\Rightarrow \pi &=& \frac{1-c}{N} \textbf{1}^{T} (I - cP)^{-1}. \\
\end{eqnarray*}

Here $\pi$ is a row vector and every other vector is a column vector. Since we are only interested in ranking we can neglect the factor $\frac{1-c}{N}$. Thus by abuse of terminology,
\begin{displaymath}
\pi^T = \textbf{1} + cP^T\pi^T.
\end{displaymath}
To estimate $\pi$, we run the following $N$ dimensional stochastic iteration. Sample $(X_n,Y_n)$ as follows: Sample $X_n$ uniformly and independently from $\{1,2,...,N\}$. Sample $Y_n$ with $P(Y_n=j|X_n=i) = p(i,j)$, independent of all other random variables realized before $n$. Update $z_n$ as follows:
\begin{equation}
z_{n+1}(i) = z_n(i) + a(n)( I\{X_{n+1}=i\}(1-z(n)) + cz_n(X_{n+1})I\{Y_{n+1} = i\}),  \label{iter1}
\end{equation}
where the step-sizes $a(n) >0$ satisfy $\sum_{n=0}^{\infty}a(n) = \infty$ and $\sum_{n=0}^{\infty}a(n)^2 < \infty$. Hence $z_n(i)$ is updated only if $X_{n+1}$, $Y_{n+1}$ or both are i. 
We can write (\ref{iter1}) as follows:
\begin{eqnarray*}
\lefteqn{z_{n+1}(i) }\\
&=& z_n(i) + a(n)\large(I\{X_{n+1}=i\}(1-z(n)) + cz_n(X_{n+1})p(X_{n+1},i) + M_{n+1}(i)\large), \\
\end{eqnarray*}
where $M_{n+1}:= cz_n(X_{n+1})I\{Y_{n+1} = i\} -  cz_n(X_{n+1})p(X_{n+1},i)$ is a martingale difference sequence w.r.t.\  ${\sigma}(X_m, Y_m, m \leq n; X_{n+1})$. By Theorem 2, p. 81, \cite{Borkar}, the ODE corresponding to the iteration is
\begin{eqnarray*}
\dot{z}(i) &=& \frac1N\large(1 + \sum_{j=1}^{N}c{z}(j)p(j,i) - z(i)\large). \\
\end{eqnarray*}
In vector form,
\begin{eqnarray*}
\dot{z} &=& \frac{1}N (\textbf{1} + cP^{T}z - z).  \\
\end{eqnarray*}
Since the constant $1/N$ doesn't affect the asymptotic behavior, we consider
\begin{eqnarray}
\dot{z} &=& (\textbf{1}+ cP^{T}z - z) =: h(z).\label{ode1}
\end{eqnarray}

 Define $h_{\infty}(z) := \lim_{a \uparrow \infty}\frac{h(az)}{a} = cP^Tz - z $.
It is easy to see that $\frac{h(az)}{a} \rightarrow h_{\infty}(z)$ uniformly on $R^N$. \\

\textbf{Theorem 1:} Under the above assumptions, $z(t) \ra z^*$ a.s., where $z^*$ is the unique solution to $h(z^*) = 0$.\\

\textbf{Proof: } Define  ${V_p}(z(t)) :=  \|z(t) - z^*\|_{p}$, $p \in [1,\infty)$.  As in the proof of Theorem 2, p.\ 126, \cite{Borkar}, for $1<p<{\infty}$,
\begin{displaymath}
\dot{V_p}(z(t)) \le \| cP^{T}(z(t) -z^*)\|_{p}-\|z(t)-z^*\|_{p}.
\end{displaymath}
Integrating,
\begin{displaymath}
{V_p}(z(t))-{V_p}(z(s)) \le \int_s^t\left(\| cP^{T}(z(r) -z^*)\|_{p}-\|z(r) -z^*\|_{p}\right) dr.
\end{displaymath}
Letting $p \downarrow1$,
\begin{eqnarray*}
{V_1}(z(t))-{V_1}(z(s)) &\le& \int_s^t\| cP^{T}(z(r) -z^*)\|_{1}-\|z(r) -z^*\|_{1} dr, \\
&\le& -\int_s^t (1-c)\| z(r) -z^*\|_{1}dr ,\\
&\le& 0,
\end{eqnarray*}
with equality iff $z(t)=z^*$. We similarly get that ${V_1}(z(t)) :=  \|z(t)\|_{1}$ is a Lyapunov function for the scaled o.d.e $\dot{z}(t) = h_{\infty}(z(t))$ which has the origin as its globally asymptotically stable asymptotic equilibrium. 
By Theorem 9, p. 75, $\cite{Borkar},\ \sup_n\|z(n)\| < {\infty}$ a.s. In turn (\ref{ode1}) has $z^*$ as its globally stable asymptotic equilibrium with $V(z(t)) = \|z(t)-z^*\|_{1}$ as its Lyapunov function. The claim follows from Theorem 7 and Corollary 8, p.\ 74, \cite{Borkar}. \hfill $\Box$

\section{Remarks}

\begin{enumerate}

\item We first look at sample complexity of the stochastic iteration. We mainly use section 4.2 of \cite{Borkar} to derive sample complexity estimates.\\

Let $1 \le m < M < N$. Let $z^*$ denote the  stationary distribution. Without loss of generality (by relabeling if necessary), let  $z^*_1 \ge z^*_2 \ge ....\ge z^*_N$, i.e., the components of $z^*$ are numbered in accordance with their ranking. We shall consider as our objective the event that the top $m$ ranks of $z^*$ fall within the top $M$ ranks of the output of our algorithm when stopped, for a prescribed pair $m < M$. This is a natural criterion for ranking problems, which are an instance of `ordinal optimization' \cite{Ho}. To avoid pathologies, we assume that $z^*_m > z^*_M$. We shall derive an estimate for the number of iterates needed to achieve our aim with `high' probability. Let $C := \{ z \in {\Delta}^N : \text{ if } z_{l_1} \ge z_{l_2} \ge ....\ge z_{l_N} \text{ then } z_i \ge z_{l_M}, 1\le i \le m\}$, where ${\Delta}^N$ is the N-dimensional probability simplex. Thus $C$ consists of all distributions such that the top $m$ indices of $z^*$ are in the top $M$ indices of the given distribution. \\

Let ${\Phi}_T$  be the time-$T$ flow-map associated with the differential equation, where $T > 0$. Thus,
\begin{displaymath}
{\Phi}_T(z) = e^{\frac{cP^T-I}{N}T}(z - (cP^T-I)^{-1}\textbf{1})- (cP^T-I)^{-1}\textbf
{1},
\end{displaymath}
with ${\Phi}_T(z^*)=z^*$. Define
\begin{displaymath}
C^* := \{z \in C : \|z - z^*\|_1 \leq \min_{z' \in \partial C} \|z' - z^* \|_{1}\},
\end{displaymath}
and for $\epsilon > 0$,
\begin{displaymath}
C^{\epsilon} := \{x : \inf_{y \in C}\|x - y\|_1 < \epsilon\}.
\end{displaymath}

Then
\begin{eqnarray*}
&& \min _{z \in \Delta^N-C}\Big[\|z-z^*\|_1 - \|{\Phi}_T(z)-z^*\|_1\Big] \\
&=& \min _{z \in \Delta^N-C}\Big[\|z-z^*\|_1 - \|e^{\frac{cP^T-I}{N}T}(z-z^*)\|_1\Big] \\
&\ge&\min _{z \in \Delta^N-C}\Big[\|z-z^*\|_1 - \|e^{\frac{cP^T-I}{N}T}\|_1\|(z-z^*)\|_1\Big] \\
&=& (1-\|e^{\frac{cP^T-I}{N}T}\|_1)\min _{z \in \Delta^N-C}\|z-z^*\|_1  \\
&=&  (1-\|e^{\frac{cP^T-I}{N}T}\|_1)\kappa \\
\end{eqnarray*}
where $\kappa := \min _{z \in \Delta^N-C}\|z-z^*\|_1$ and $\|A\|_1$ for a matrix $A$ is its induced matrix norm. We argue that $\|e^{\frac{cP^T-I}{N}T}\|_1 < 1$. To see this, view $Q = cP - I$ as the rate matrix of a continuous time Markov chain killed at rate $1 - c$. Then $e^{\frac{cP-I}{N}T}$ is its transition probability matrix after time $\frac{T}{N}$, whose row sums will be uniformly bounded away from $1$. The claim follows. Let $\gamma > 0$ and pick $T > 0$ such that
\begin{displaymath}
\gamma \geq \min _{z \in \Delta^N-C}[\|z-z^*\|_1 - \|{\Phi}_T(z)-z^*\|_1].
\end{displaymath}
Since $\max_{z \in \Delta^N} \|z-z^*\|_{1} = 2$,
  \begin{displaymath}
 \frac{ \max_{z \in \Delta^N} \|z-z^*\|_{1}}{\gamma/2}\times(T+1) \leq \tau := \frac{4}{(1-\|e^{\frac{cP^T-I}{N}T}\|_1)\kappa}\times(T+1).
 \end{displaymath}
Let $n_0 := \min\{n \geq 0 : \sum_{m=0}^na(m) \geq \tau\}$. Also, let $\mathcal{N}_{\eta}(S) := \{z : \inf_{y \in S}\|z - y\|_2 \leq \eta\}$ denote the open $\eta$-neighborhood w.r.t.\ $\| \ \cdot \ \|_2$ norm of a generic set $S$.
Set $\delta := \frac{\gamma}{2\sqrt{N}}$. Then $\|x - y\|_2 < \delta \Longrightarrow \|x - y \|_1 < \frac{\gamma}{2}$.
 Arguing as in Corollary 14, p.\ 43 of \cite{Borkar}, we have
\begin{displaymath}
P(z_n \in N_{\delta}(C^{\frac{\gamma}{2}}) \  \forall n \ge n_0 + k) \ge 1 - 2Ne^{-\frac{K\delta^2}{N\sum_{m = k}^{\infty}a(m)^2}} = 1 - o(\sum_{m = k}^{\infty}a(m)^2),
\end{displaymath}
where $K > 0$ is a suitable constant.  (In \textit{ibid.}, replace $H^{\epsilon}$ by $C^*$ and $\Delta$ by $\gamma$.)

\item Note that at each time $n$, we can generate more than one, say $m$  pairs $(X^i_n, Y^i_n), 1 \leq i \leq m$, which are independent, each distributed as $(X_n, Y_n)$ above, and change the iteration to:
\begin{eqnarray*}
z_{n+1}(i) &=& z_n(i) + a(n)( I\{i \in \{X^j_{n+1}, 1 \leq j \leq m\}\}(1-z(n)) \\
&& \ + \ c\sum_{j=1}^{m}z_n(X^{j}_{n+1})I\{Y^{j}_{n+1}=i\}).
\end{eqnarray*}
That is, we update several components at once. This will speed up convergence at the expense of increased per iterate computation.
\end{enumerate}

\section{Numerical Experiments}
In this section we simulate the algorithm for different number of nodes. The results for the cases when the number of nodes are 50, 200 and 500 are plotted in Figure 1, Figure 2 and Figure 3 resectively. The dotted line indicates the distance between $z^*$ and $z_n$ w.r.t.\ $n$. The solid line indicates the percentage of top 5 indices of $z^*$ that do not feature in the top 10 indices of $z_n$. Figure 4, Figure 5 and Figure 6 further show (for 200 nodes) that the number of iterations required to achieve this objective varies inversely with variance of $z^*$.   \\

\centerline{\includegraphics[scale=0.8]{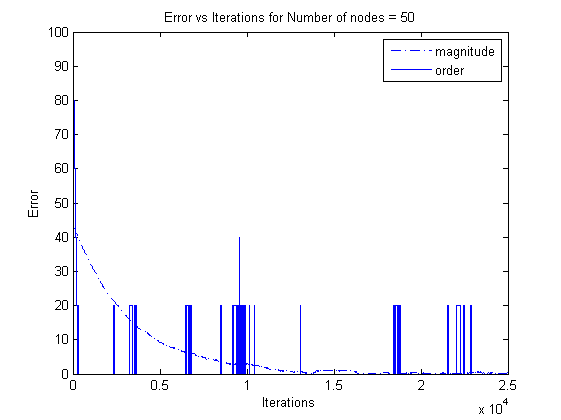}}
\centerline{Figure 1: varaince of $z^*$=47.1641} 

\vspace{3mm}

\centerline{\includegraphics[scale=0.8]{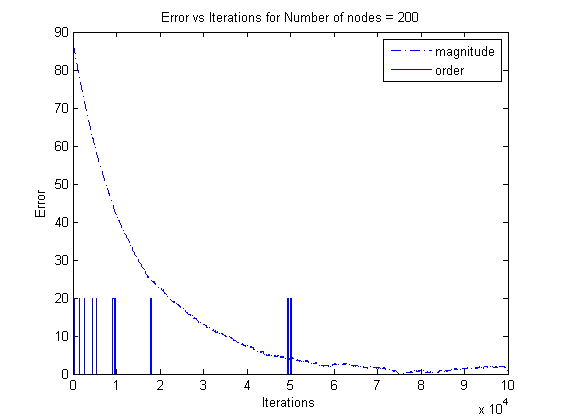}}
\centerline{Figure 2: variance of $z^*$=277.3392} 

\vspace{5mm}

\centerline{\includegraphics[scale=0.8]{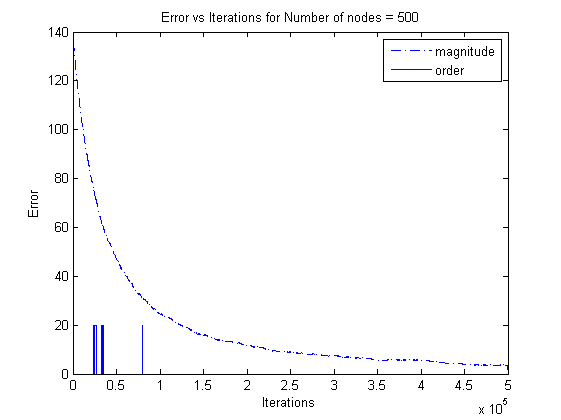}}
\centerline{Figure 3: variance of $z^*$ = 743.4651}

\vspace{5mm}

\centerline{\includegraphics[scale=0.8]{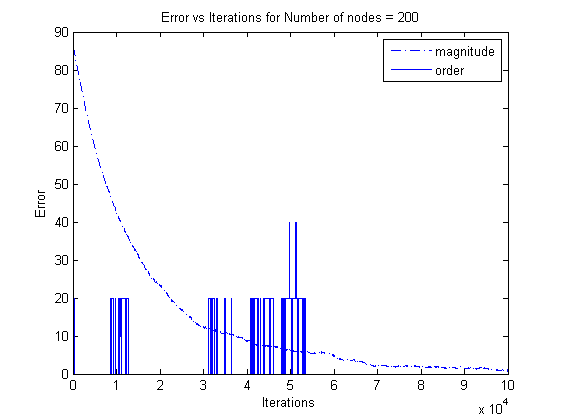}}
\centerline{Figure 4: varaince of $z^*$=259.6187} 

\vspace{5mm}

\centerline{\includegraphics[scale=0.8]{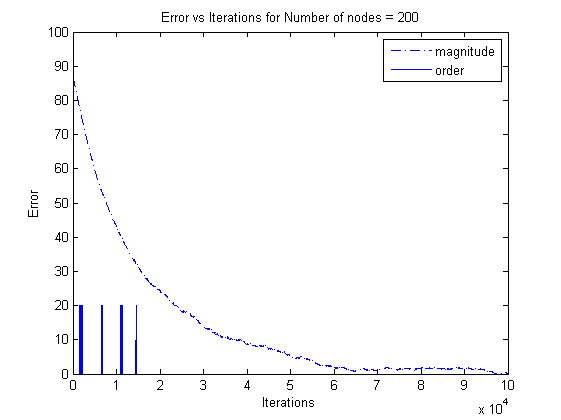}}
\centerline{Figure 5: variance of $z^*$=335.6385} 

\vspace{5mm}

\centerline{\includegraphics[scale=0.8]{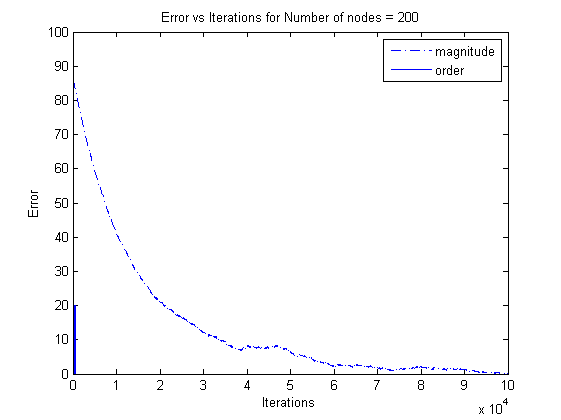}}
\centerline{Figure 6: variance of $z^*$ = 365.0774}

\vspace{5mm}

\section{Conclusions}

In conclusion, we highlight some of the important features of the above scheme, which are facilitated by the reinforcement learning framework.

\begin{enumerate}
\item As already mentioned, the scheme does not depend on an a priori model for transition probabilities, but is completely data-driven in this aspect.

\item We use `split sampling' introduced in \cite{ISI} for reinforcement learning, sampling pairs $(X_n, Y_n)$ with the desired conditional law for $Y_n$ given $X_n$, but with uniform sampling for $\{X_n\}$. This is a departure from classical reinforcement learning, where one runs a single Markov chain $\{X_n\}$ according to $\hat{P}$ and $Y_n = X_{n+1}$.

\item Since we are iterating over probability vectors as they evolve under a transition matrix, the scheme requires \textit{left-multiplication} by row vectors thereof. This is different from usual reinforcement learning schemes, which involve averaging with respect to the transition probabilities, i.e., \textit{right-multiplication} by a column vector. We have worked around this difficulty by modifying the update rule. In classical reinforcement learning algorithms based on a simulated Markov chain $\{X_n\}$, one updates the $X_n$th component at time $n$, i.e., the $i$th component gets updated only when $X_n = i$. In the above scheme, the $i$th component gets updated both when $X_{n+1} =
    i$ and when $Y_{n+1} = i$, albeit in different ways. This is another novel feature of the present scheme.

\end{enumerate}

\end{document}